\pdfoutput=1

\documentclass[11pt]{article}

\usepackage[]{EMNLP2023}

\usepackage{times}
\usepackage{latexsym}

\usepackage[T1]{fontenc}

\usepackage[utf8]{inputenc}

\usepackage{microtype}

\usepackage{inconsolata}

\usepackage{multirow}
\usepackage{graphicx}
\usepackage{booktabs}   
\usepackage{adjustbox}
\usepackage{subfigure}
\usepackage{tabularx}
\usepackage{amsmath}
\usepackage[frozencache, cachedir=.]{minted}
\usepackage{enumitem}
\usepackage{xspace}
\usepackage{algorithm, algpseudocode}
\usepackage{amssymb}
\usepackage{booktabs} 
\usepackage{xcolor} 
\usepackage{tabularx} 

\usepackage{verbatimbox} 
\usepackage{fancyvrb}
\usepackage{listings}
\usepackage{scalefnt}
\usepackage[T1]{fontenc}

\let\emptyset\varnothing

\def \name{\textsc{SelectLLM}\xspace}
\definecolor{Gray2}{gray}{0.7}

\title{
\name{}: Can LLMs Select Important Instructions to Annotate?
}

\author{
        Ritik Sachin Parkar$^{\ddagger,}$\thanks{~~Equal Contribution.} \quad
        Jaehyung Kim$^{\diamondsuit, *}$ \quad
	Jong Inn Park$^{\ddagger}$ \quad
	Dongyeop Kang$^{\ddagger}$ 
 \\
	$^\ddagger$University of Minnesota, \quad $^\diamondsuit$Carnegie Mellon University \\
	{\tt \{parka438, park2838, dongyeop\}@umn.edu jaehyun4@andrew.cmu.edu} 
}

\begin{document}
\maketitle
\begin{abstract}
Instruction tuning benefits from large and diverse datasets; however, creating such datasets involves a high cost of human labeling. 
While synthetic datasets generated by large language models (LLMs) have partly solved this issue, they often contain low-quality data.
One effective solution is selectively annotating unlabelled instructions, especially given the relative ease of acquiring unlabeled instructions or texts from various sources. 
However, how to select unlabelled instructions is not well-explored, especially in the context of LLMs. 
Therefore, we introduce \name{}, an alternative framework that leverages the capabilities of LLMs to select unlabeled instructions more effectively.
Specifically, \name{} consists of two key steps: Coreset-based clustering of unlabelled instructions for enlarging diversity and prompting of LLM to identify the most beneficial instructions within each cluster. 
We evaluate \name{} on AlpacaEval2 and MT-Bench, demonstrating its ability to outperform state-of-the-art methods like Alpagasus. 
In addition, we compare the performance and compatibility of \name{} with various LLMs, such as ChatGPT, LLaMA-3.1-70B, and Gemma-2-27b.
\name{}'s adaptability and robustness are further evidenced by its ability to maintain high performance across both human and synthetic datasets. All code and data are publicly available\footnote{\url{https://github.com/minnesotanlp/select-llm}}.

\end{abstract}

\section{1. Introduction}

Instruction tuning, which fine-tunes language models (LMs) to follow human instructions constructed from diverse tasks, has shown impressive generalization performance on unseen tasks \citep{wei2022finetuned, chung2022scaling}. 
However, creating large and diverse annotated instruction datasets is a major challenge due to the huge cost of human labeling. 
While synthetic datasets labeled by advanced large language models (LLMs) have partly addressed this issue \citep{alpaca, selfinstruct}, they often contain low-quality data, highlighting the need for more focus on dataset refinement \citep{zhou2023lima, cao2023instruction, das2024surface}. 

\begin{figure}[t!]
	\centering
	\includegraphics[width=1.0\columnwidth,clip]{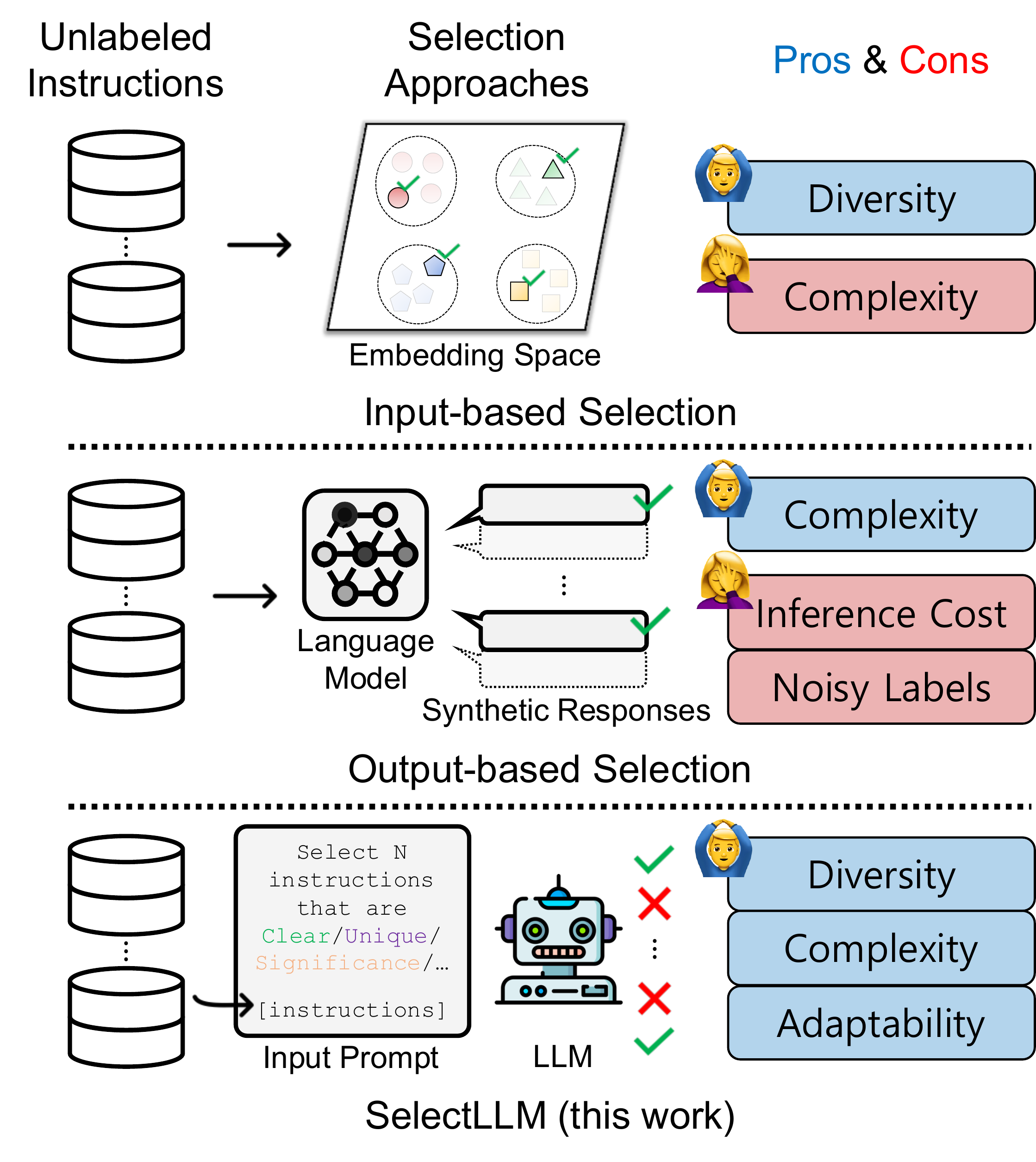}
    \vspace{-0.25in}
    \caption{Conceptual comparison between previous approaches to select instructions and \name{}. Focusing on input instructions (\textbf{top}) is unable to consider the difficulty or uncertainty of response. Output-based methods (\textbf{middle}) can suffer from the inference cost and quality issues of synthetic responses.
    \name{} (\textbf{bottom}) does not suffer from these issues by estimating the effectiveness of instructions via prompting LLMs.}
    \label{fig:figure1}
    \vspace{-0.15in}
\end{figure}

A promising approach to overcoming these challenges is to select a smaller subset of high-quality unlabelled instructions and annotate it (similar to active learning \citep{settles2009active}), as accessing unlabelled instructions from various sources \citep{instructionwild} is relatively easy.
However, existing methods for selecting such \textit{unlabeled} instructions have limitations; 
active learning algorithms that prioritize input density in the embedding space \citep{sener2018active} often fail to account for the complexity and diversity of the response (label) for the given instruction. 
Alternatively, output-based methods that assess {uncertainty} in model predictions \citep{kung2023active} or their quality \citep{chen2023alpagasus} struggle with high computational costs and the challenges from the quality of synthetic labels (Figure \ref{fig:figure1}).

To address these limitations, we investigate a crucial question, inspired by recent works demonstrating the ability of LLMs to measure the quality or relevance of the texts \citep{sun2023chatgpt, liu2023gpteval}:  \textit{Can LLMs select effective unlabelled instructions by leveraging their vast knowledge base to discern the complexity and utility of each instruction?} 

\begin{figure}[t]
	\centering
	\includegraphics[width=0.9\columnwidth,clip]{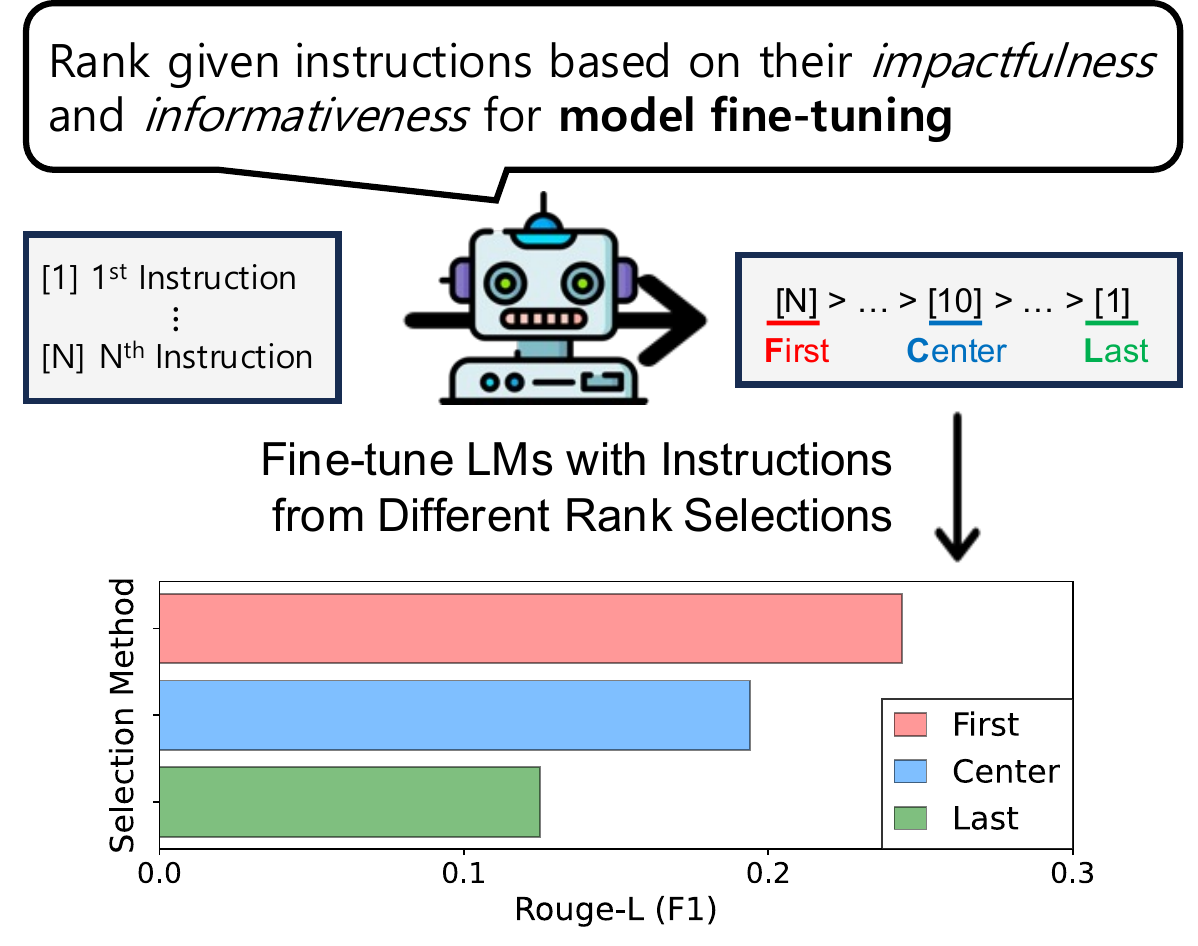}
    \caption{Experiments to verify  LLMs' capability to infer the importance of \textit{unlabelled} instructions. We prompt ChatGPT to sort the instructions based on their effectiveness for model training; then, we compare the performance of three fine-tuned LMs (LLaMA-2) on instructions with different ranks (First, center, and last). A full prompt is presented in Appendix \ref{appendix:list_prompt}.}
    \label{fig:figure_intuition}
    \vspace{-0.15in}
\end{figure}

\textbf{Contribution.}
In this work, we introduce \name{}, a novel framework that selects an effective subset of \textit{unlabelled} instructions by prompting LLM. 
At a high level, we use LLM to estimate the usefulness and impact of each instruction \textit{without corresponding labels}; through initial experiments presented in Figure \ref{fig:figure_intuition}, we verify that LLMs indeed possess such capability. 
To further improve the selection via LLMs, \name{} first divides the entire set of whole unlabeled instructions into several small subsets; 
specifically, we use equal-size K-means clustering to create each subset with diverse unlabelled instructions while preserving the overall dataset structure. 
Then, \name{} constructs an input query for each subset using a specifically designed input prompt for selecting with LLMs and forwards them to LLM to select a few instructions expected to be most helpful in fine-tuning models.
While prior work has used prompting to filter low-quality labeled instructions on an instance-by-instance basis \cite{chen2023alpagasus}, our approach is the first to explore direct selection from unlabeled instructions. Further, it is more effective as it enables batched selection.

We demonstrate the effectiveness of \name{} compared to various SoTA selection methods on the Dolly \citep{DatabricksBlog2023DollyV2} and Cleaned Alpaca \citep{alpaca} datasets. 
Our experiments reveal that \name{} consistently outperforms these baselines across varied selection sizes. 
Notably, \name{} surpasses the SoTA baseline Alpagasus \citep{chen2023alpagasus}, which requires labeled instructions, on the Dolly dataset using both MT-Bench \citep{zheng2024judging} and AlpacaEval2 \citep{dubois2024length} benchmarks, supported by human evaluation as well. We also demonstrate how \name{} offers a cost-efficient way of using paid models, such as ChatGPT, compared to Alpagasus. Additionally, we compare the compatibility and performance of open-source LLMs with \name{} relative to paid GPT models. Furthermore, the input prompt in \name{} is designed to provide flexibility for easy customization to meet specific user needs, including reducing toxicity— something not offered by previous selection methods. This work introduces a novel approach to instruction selection and paves the way for leveraging LLMs to efficiently create high-quality, tailored instruction datasets.

\section{Related Work}\label{sec:2}

\begin{figure*}[t]
    \centering
    \includegraphics[width=1.0\textwidth,trim={0cm 0cm 0cm 0cm},clip]{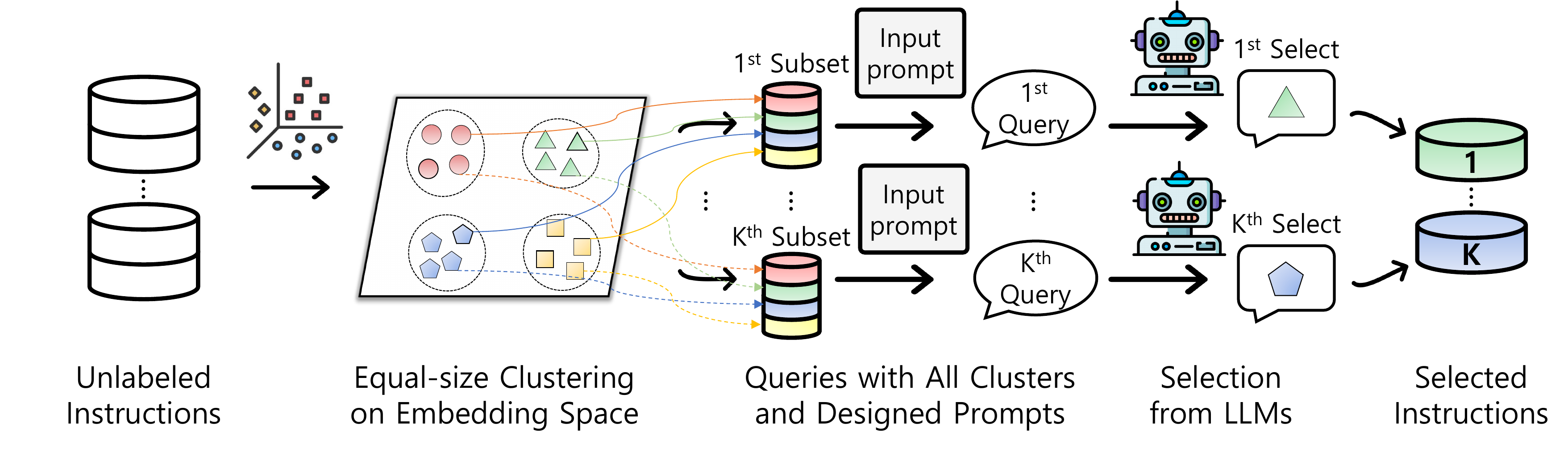}
    \vspace{-0.25in}
    \caption{Illustration of the proposed \name{}.}
    \label{fig:figure2}
    \vspace{-0.1in}
\end{figure*}

\noindent\textbf{Instruction tuning for LMs.} 
Instruction tuning \citep{wei2022finetuned}, a form of fine-tuning LLMs, has emerged as a prominent methodology to align pre-trained LMs for various tasks by describing the tasks in the common form of instructions. 
Due to its ease of implementation and remarkable generalization capabilities for unseen tasks \citep{wei2022finetuned, chung2022scaling, jang2023exploring}, it has gained substantial popularity recently. 
Constructing these instruction datasets with human annotations is a standard way \citep{DatabricksBlog2023DollyV2, sanh2021multitask, supernaturalinstructions}, but this method faces challenges in terms of variety of instructions and the total number of instances due to labeling cost \citep{selfinstruct}; 
one promising solution for this limitation is to synthesize existing datasets and create diverse, multi-task datasets with the help of LLMs such as Alpaca \citep{alpaca} and Self-instruct \citep{selfinstruct}.
However, using LLM-created data increases the risk of including low-quality examples, and it is known that removing such noise from the dataset is critical for the effective instruction tuning of LLMs \citep{zhou2023lima, cao2023instruction, das2024surface}; 
therefore, in this work, we explore an alternative way to use LLMs to construct a high-quality instruction dataset, by using LLMs to select unlabeled instructions.

\noindent\textbf{Sample selection for efficient instruction tuning.} 
Expanding instruction datasets has its own set of challenges, including the need for extensive resources, time, human annotation, and the prevalence of redundant data.
A common solution involves human intervention, such as the approach by \citet{zhou2023lima}, which manually annotates high-quality instructions after filtering out low-quality data. 
However, there are huge costs associated with such procedures done by human labor. 
An effective solution is to gather a large pool of unlabeled instructions and then selectively annotate the most useful ones, similar to the popular task of active learning \citep{settles2009active, sener2018active}. 
Here, the most important thing is \textit{selection criteria}.

One line of work focuses on \textit{density} to include \textit{diverse} instructions; for example, \citet{chen2023maybe} map the unlabelled instructions into an embedding space and use K-means clustering \citep{hartigan1979algorithm} and the K-center greedy algorithm \citep{sener2018active} for the selection.
The idea of clustering to select instruction data is also adopted in a vision-language domain \citep{wei2023instructiongpt}.  
Another line of work focuses on \textit{uncertainty} (or \textit{difficulty}) of instructions measured with LLMs' outputs; \citet{kung2023active} measure uncertainty by observing how LLM-generated responses vary with changes in the input instructions. 
Unlike these approaches, we directly prompt LLMs with unlabelled instructions, to select the few examples expected to help train the model; from LLM's capability of reasoning and generating useful responses, we assume that they could infer the impactfulness of the unlabelled instructions. Meanwhile, recent studies by \citep{chen2023alpagasus} and \citep{li2024shot} also prompt LLMs to construct high-quality instruction datasets. However, our work differs as we focus on selecting unlabelled instructions while they focus on filtering out low-quality labeled ones.

\section{\name{}: Select Important Unlabeled Instructions Using LLMs}\label{sec:3}

\subsection{Preliminary}
\label{sec:3.1}
We first describe the problem setup of our interest.
Let $\mathcal{X}=\{x_{i}\}_{i=1}^{M}$ denote the given unlabeled dataset, where $x_{i}$ represents $i$th unlabeled instruction and $M$ is the total number of instructions.  
Then, our goal is to select $N$ most effective unlabeled instructions from $\mathcal{X}$ which will be labeled by human annotators, to fine-tune a target large language model (LLM) $f_{\theta}$, \textit{e.g.}, LLaMA-2 \citep{touvron2023llama}, and make it be generalized for various instructions.
Formally, we select $N (<M)$ instructions under a selection criteria $s(x) \in [0,1]$ (\textit{i.e.}, $s(x)=1$ indicates the selection of $x$):  
\begin{equation}
    \mathcal{X}(s, N)=\{x \in \mathcal{X}|s(x)=1, \sum\nolimits_{x \in \mathcal{X}} s(x)=N\} 
\end{equation}
Then, the selected instruction $x \in \mathcal{X}(s, N)$ is labeled with a corresponding label $y$ by human annotators, and it results in the annotated instruction dataset $\mathcal{D}$, where $\mathcal{D}=\{(x_{j},y_{j})\}_{j=1}^{N}$. 
Therefore, the performance of LLM $f_{\theta}$ fine-tuned on $\mathcal{X}(s,N)$ significantly varies depending on which selection criteria $s(x)$ is considered. 
While various selection criteria have been explored under tasks like active learning (see Sec \ref{sec:2}), this direction is less explored under the paradigm of instruction tuning.

\begin{figure}[t!]
    \centering
    \begin{lstlisting}
The following are @{N}@ candidate instructions that describe a task, each indicated by a number identifier [].
$
[1]
### Instruction: {Example #1 Instruction}
### Input: {Example #1 Input}
.
.
.
[N]
### Instruction: {Example #N Instruction}
### Input: {Example #N Input}
$
Examine the provided list of @{N}@ instructions, each uniquely identified by a number in brackets []. 

Your task is to select @{num}@ instructions that excel in various aspects.

Look for instructions that are clear and relevant, exhibit a high level of complexity and detail, represent a diverse range of scenarios and contexts, offer significant instructional value and potential learning gain, and present unique challenges and specificity.

These selected instructions should ideally be the most beneficial for model fine-tuning after being annotated by human annotators.
       
Present your selections using the format []. e.g., [1,2] or [2,3].

The most impactful @{num}@ instructions (only identifiers) are:
    
    \end{lstlisting}
    \caption{Designed input prompt of \name{}.}
    \label{fig:prompt}
    \vspace{-0.1in}
\end{figure}

\subsection{Selection via prompting LLMs}
\label{sec:3.2}
For the selection criteria $s(x)$, \name{} proposes to use LLMs with a properly designed prompt without using ground truths or generated labels. 
Our high-level intuition is that LLMs can infer the potential impact of each instruction by only reading the instruction; as shown in Figure \ref{fig:figure_intuition}, we observed that the recent LLM, \textit{e.g.}, ChatGPT, could estimate the effectiveness of each instruction for model training (\textit{e.g.}, mistral-7b-v0.3), even without the corresponding labels. 
To further improve the effectiveness of selection via LLMs, we carefully designed the input prompt to incorporate several important perspectives for instruction tuning, and it is presented in Figure \ref{fig:prompt}.
Formally, this process could be described as follows: we first assume that the dataset $\mathcal{X}$ is divided into $K$ non-overlapped subsets, \textit{i.e.}, $\mathcal{X}=\bigcup_{k=1}^{K} \mathcal{X}_{k}$.
Then, we construct input query $q_{k}$ using the designed prompt $p_{\tt sel}$ and $\mathcal{X}_{k}$, and forward it to \texttt{LLM} to select $\widetilde{N}=\lfloor N/K \rfloor$ examples:
\begin{equation}
      S_{k} = \texttt{LLM}\big(p_{\tt sel}(q_{k}, \widetilde{N})\big) 
\end{equation}
where $S_{k}=\{s(x)\in[0,1]| x\in\mathcal{X}_{k}, \sum\nolimits_{x\in\mathcal{X}_{k}} s(x) =\widetilde{N}\}$. 

\subsection{Composing query of LLMs via clustering}\label{sec:3.3}
To further improve the effectiveness of using LLMs for selection, we carefully design how to divide the entire dataset into several subsets which would be used to construct input queries, based on the equal-size clustering method.
Here, our high-level idea is composing the subsets that maximize the diversity among the instructions while maintaining the global structure of the dataset.  
Specifically, we first extract the embeddings of the instructions in $\mathcal{X}$, using the pre-trained sentence encoder $g_{\phi}$ such as Sentence-BERT \citep{reimers2019sentence}.  
Then, we conduct K-means clustering \citep{hartigan1979algorithm} on these embeddings, and calculate $D \in \mathbb{R}^{N \times K}$, the distance of all instances in $\mathcal{X}$ to $K$ cluster centers $c_{1}, \cdots, c_{K}$.
Based on the distances, we assigned each instance $x$ among $[1,K]$, by iteratively taking the one with the shorted distance to the cluster center among the remaining instances, to guarantee equal sizes for each $k$. 

Overall, the selection procedure of \name{} is as follows: (1) construct input queries by separating the entire dataset into multiple subsets of diverse instructions. Then, (2) feed these queries into a LLM and get the selected indices. The formal presentation of these procedures is presented in Algorithms \ref{alg:main} and \ref{alg:kmeans} in Appendix.

\section{Experiments}

\begin{table*}[t]
    \caption{Experimental results on Dolly \citep{DatabricksBlog2023DollyV2}. Rouge-L (F1) and Cosine similarity of generated responses from fine-tuned LLaMA-2 by different numbers of examples are compared. The best and second best scores are highlighted in \textbf{bold} and \underline{underline}, respectively.}
    \vspace{-0.1in}
	\begin{center}
	\begin{adjustbox}{width=0.85\linewidth}
	\begin{tabular}{r|ccc|ccc|ccc}
 		\toprule
		 & \multicolumn{3}{c}{Rouge-L (F1)} & \multicolumn{3}{c}{Cosine Similarity} & \multicolumn{2}{c}{Avg Across Sizes} \\ 
		Methods & $1k$ & $2k$ & $3k$ & $1k$ & $2k$ & $3k$ 
  & Avg Rouge & Avg Cosine \\ \midrule
  Length\textsubscript{short}
  & 0.073 & 0.109 & 0.130
  & 0.192 & 0.265 & 0.336 
  & 0.104 & 0.264  \\ 
  Perplexity
  & 0.158 & 0.183 & 0.192 
  & 0.402 & 0.433 & 0.453 
  & 0.178 & 0.429 \\
  CBS\textsubscript{sbert}
  & 0.147 & 0.200 & 0.216 
  & 0.359 & 0.473 & 0.512 
  & 0.188 & 0.448 \\
  Length\textsubscript{long}
  & 0.256 & 0.247 & 0.238
  & 0.641 & 0.626 & 0.611 
  & 0.247 & 0.626 \\ 
  CBS\textsubscript{instr}
  & 0.258 & 0.255 & 0.255 
  & 0.617 & 0.638 & 0.632 
  & 0.256 & 0.629 \\
    Random 
  & 0.239 & 0.264 & 0.278 
  & 0.589 & 0.644 & 0.650 
  & 0.260 & 0.628 \\
  Diversity
  & 0.237 & 0.275 & 0.282 
  & 0.582 & 0.650 & 0.666 
  & 0.265 & 0.633 \\
  OpenEnd
  & 0.258 & 0.271 & \underline{0.282} 
  & 0.627 & 0.641 & \underline{0.669}
  & 0.270 & 0.646 \\
  Coreset
  & \underline{0.271} & \underline{0.281} & 0.279 
  & \underline{0.649} & \underline{0.662} & 0.659 
  & \underline{0.277} & \underline{0.657} \\ \midrule
  Ours
  & \textbf{0.278} & \textbf{0.288} & \textbf{0.289} 
  & \textbf{0.668} & \textbf{0.680} & \textbf{0.686} 
  & \textbf{0.285} &\textbf{0.678} \\ 
 		\bottomrule
	\end{tabular}
    \end{adjustbox}
    \end{center}
    \vspace{-0.1in}
    \label{table:main_dolly}
\end{table*}

\subsection{Setups}

\noindent\textbf{Datasets and metrics.} We use labeled datasets without using their responses to test our hypothesis. We utilize one human-generated dataset, Dolly \citep{DatabricksBlog2023DollyV2}, a combined effort of several Databricks employees, and one machine-generated dataset, Cleaned Alpaca, based on \citep{alpaca} but cleaned to fix errors in the original input prompts with responses generated by GPT-4. For performance evaluation, we assess the similarity between inferred and actual texts using Rouge scores and cosine similarity with various baselines. Additionally, we conduct MT-Bench and AlpacaEval2 evaluations on \name{}, random selection, and a SoTA baseline, Alpagasus. MT-Bench assesses multi-turn conversation coherence, while AlpacaEval2 evaluates single-turn task performance.

\noindent\textbf{Baselines.} We consider several baselines for comparison with our algorithm as follows: 
\textit{(1) Random}: selecting instances from the unlabeled dataset purely randomly. 
\textit{(2) Length}: Considers the length of input instruction, focusing on both longer and shorter ones to evaluate their impact (Length\textsubscript{long} and Length\textsubscript{short}). 
\textit{(3) Cluster-Based Selection (CBS)} \citep{chen2023maybe}: transforming instructions into embedding space, clustering them with HDBSCAN \citep{campello2013hdbscan}, and selecting samples using the K-Center-Greedy algorithm. We consider two different embedding spaces with Sentence-BERT \citep{reimers2019sentence} and InstructOR \citep{su2022one}, and denote them as CBS\textsubscript{sbert} and CBS\textsubscript{instr}, respectively. 
\textit{(4) Perplexity} \citep{marion2023more}: selecting samples based on low per-token perplexity, indicating high model certainty and fluency. 
\textit{(5) Diversity} \citep{selfinstruct}: for each instruction in the dataset, Rouge score is computed against a randomly selected subset comprising $n$ samples ($n \ll M$). Then, we select $k$ samples that exhibit the minimum Rouge scores. 
\textit{(6) Open-Endedness (OpenEnd)} \citep{li2023symbolic}: generating three inferences per prompt, counting unique bigrams, and selecting samples with the greatest variety of bigrams. 
\textit{(7) Coreset} \citep{sener2018active}: Similar to CBS, transforming instructions into embedding space with Sentence-BERT, then selecting samples with K-Center-Greedy algorithm \citep{sener2018active}. 
\textit{(8) Alpagasus} \citep{chen2023alpagasus}: Scoring each data point with an auto-grader like ChatGPT based on dimensions such as helpfulness or accuracy and filtering out low-scored instances.

\begin{table*}[t]
    \caption{Experimental results on Cleaned Alpaca \citep{DatabricksBlog2023DollyV2}. Rouge-L (F1) and Cosine similarity of generated responses from fine-tuned LLaMA-2 by different numbers of examples are compared. The best and second best scores are highlighted in \textbf{bold} and \underline{underline}, respectively.}
    \vspace{-0.1in}
	\begin{center}
	\begin{adjustbox}{width=0.85\linewidth}
	\begin{tabular}{r|ccc|ccc|ccc}
 		\toprule
		 & \multicolumn{3}{c}{Rouge-L (F1)} & \multicolumn{3}{c}{Cosine Similarity} & \multicolumn{2}{c}{Avg Across Sizes} \\ 
		Methods & $1k$ & $2k$ & $3k$ & $1k$ & $2k$ & $3k$ 
  & Avg Rouge & Avg Cosine \\ \midrule
  Length\textsubscript{short}
  & 0.219 & 0.263 & 0.261
  & 0.519 & 0.632 & 0.625 
  & 0.248 & 0.592  \\ 
  Perplexity
  & 0.264 & 0.278 & 0.272 
  & 0.628 & 0.646 & 0.640 
  & 0.271 & 0.638 \\
  CBS\textsubscript{sbert}
  & 0.254 & 0.264 & 0.288 
  & 0.598 & 0.618 & 0.665 
  & 0.269 & 0.627 \\
  Length\textsubscript{long}
  & 0.228 & 0.266 & 0.297
  & 0.550 & 0.618 & 0.683 
  & 0.264 & 0.617 \\ 
  CBS\textsubscript{instr}
  & 0.257 & 0.280  & 0.292  
  & 0.610 & 0.655  & 0.673  
  & 0.276 & 0.646  \\
  Random 
  & \textbf{0.281} & \underline{0.281} & 0.271 
  & \underline{0.653} & 0.662 & 0.656 
  & 0.278 & 0.657 \\
  Diversity
  & 0.268 & \textbf{0.286} & 0.297 
  & 0.650 & \textbf{0.673} & 0.690 
  & \underline{0.284} & \underline{0.671} \\
  OpenEnd
  & 0.250 & 0.247 & 0.276 
  & 0.616 & 0.601 & 0.645
  & 0.258 & 0.621 \\
  Coreset
  & \underline{0.277} & 0.267 & \underline{0.299}
  & 0.650 & 0.635 & \underline{0.695} 
  & 0.281 & 0.660 \\ \midrule
  Ours
  & 0.276 & 0.277 & \textbf{0.301} 
  & \textbf{0.661} & \underline{0.667} & \textbf{0.707}
  & \textbf{0.285} &\textbf{0.678} \\ 
 		\bottomrule
	\end{tabular}
    \end{adjustbox}
    \end{center}
    \vspace{-0.1in}
    \label{table:main_alpaca}
\end{table*}
\begin{table}[t]
    \caption{MT-Bench and AlpacaEval2 evaluation results of fine-tuned Mistral-7B-inst-v0.3 on $3k$ samples of Dolly dataset selected with different methods. The MT-Bench scores range from 1 to 10, while AlpacaEval2 metrics include Length-Controlled (LC) win rates and raw win rates (WR) as percentages. The highest scores are \textbf{bolded}.}
    \vspace{-0.1in}
	\begin{center}
	\begin{adjustbox}{width=1\linewidth}
	\begin{tabular}{l|c|c|c>{\centering\arraybackslash}p{2em}|>{\centering\arraybackslash}p{2em}}
            \toprule
            \multirow{2}{*}{Model} & MT-Bench & \multicolumn{2}{c}{AlpacaEval2 (\%)} \\
            & (1-10) & LC & WR \\
            \midrule
            Mistral-7B (vanilla) &  2.33 & - & - \\
            Random & 5.44 & 4.00 & 1.93 \\
            Alpagasus & 5.69 & 4.76 & 2.48 \\ \midrule
            SLLM GPT 3.5 & \textbf{6.12} & \textbf{6.72} & \textbf{3.29} \\
            SLLM Llama & 3.36 & 4.75 & 2.3 \\
            SLLM Gemma & 5.00 & 4.32 & 2.11 \\
            \bottomrule
    \end{tabular}
    \end{adjustbox}
    \end{center}
    \vspace{-0.1in}
    \label{table:main_benchmark}
\end{table}

\noindent\textbf{Implementation Details.} For the Rouge and cosine similarity evaluations on the labeled datasets, we allocated 1k samples from each for testing, leaving 14k and 51k samples in the Dolly and Cleaned Alpaca datasets, respectively, for training. Subsets of 1k to 3k samples were selected from each train set using different sampling algorithms. The LLaMA-2 (7B) model was fine-tuned using QLoRa \citep{touvron2023llama} to optimize memory usage during training. To ensure robustness, we conducted experiments with three random seeds and averaged the results. In the second evaluation involving MT-Bench and AlpacaEval2, we used all 15k and 52k samples from Dolly and Cleaned Alpaca, respectively, fine-tuning the Mistral-7b-v0.3 \citep{jiang2023mistral7b} with QLoRa. For instruction selection, we applied \name{}, alongside ChatGPT (\texttt{gpt-3.5-turbo-0125}), Llama3.1-70B-IT with 8 bits Quantization, and Gemma 2-27B-IT \citep{gemmateam2024gemmaopenmodelsbased}. Detailed training procedures are documented in the Appendix.

\subsection{Main Results}
In this section, we present our main experimental results. 
We first evaluate \name{} and the baseline selection methods by fine-tuning LLaMA-2 on the selected samples and then evaluate the generated responses for the test instructions using two popular metrics Rouge-L (F1) and Cosine Similarity. 
To provide detailed insight, we measure the performance variations across different numbers of selections (1k, 2k, and 3k) and the average performance across these sizes. 
The results for Dolly are in Table \ref{table:main_dolly} and for Alpaca in Table \ref{table:main_alpaca}, respectively. 
In addition, we compare our method with various LLMs (Open-sourced and Paid) against the current SoTA Alpagasus baseline on the MT-Bench and AlpacaEval2 benchmarks in Table \ref{table:main_benchmark}.
Consequently, our analysis leads to the following nuanced observations:

\noindent\textbf{Dominant performance of \name{}.} \textit{SelectLLM} consistently outperforms other methods in the Dolly dataset, maintaining a lead with an average improvement of 2.6\% in Rouge Score and 3\% in Cosine Similarity across all sample sizes. 
This highlights \name{}'s adaptability and effectiveness in processing human-generated data. 
In the Cleaned Alpaca dataset, \name{} shows its strength, particularly at the 1k and 3k sample sizes, outperforming others on the cosine similarity metric. 
While its performance at the 2k size is slightly lower, the overall trend underscores its reliability across various data volumes.

\noindent\textbf{Consistent effectiveness across datasets.} 
\name{} exhibits consistent improvements in both human and synthetic datasets.  
In contrast, other methods like Coreset, Diversity, and Length\textsubscript{long} show fluctuating performances depending on the dataset and sample size. 
For example, Coreset varies notably with sample size, while Diversity and Length\textsubscript{long} excel in the Cleaned Alpaca dataset but falter in the Dolly dataset. 
OpenEndedness performs better in Dolly but shows decreased effectiveness in Cleaned Alpaca. 
This uniformity across sample sizes sets \name{} apart from other baselines and demonstrates its broad applicability.

\noindent\textbf{Improvements on MT-bench and AlpacaEval2.}
\name{} with GPT 3.5 outperforms other baselines, including Alpagasus, on both single-turn and multi-turn instruction following capabilities on Dolly. 
For example, compared to Alpagasus, \name{} exhibits 1.96\% improvement in length-controlled win rate on AlpacaEval2 and 7.56\% relative improvement on MT-bench, respectively.
This result is remarkable, especially considering the efficiency of \name{}; \name{} does not require the output labels for the selection, unlike Alpagasus, and consequently, it consumes a much smaller API cost for the selection, \textit{e.g.}, \name{}: 2.82\$ v.s. Alpagasus: 23.76\$ for selecting 3k samples among 52k samples of Cleaned Alpaca. The clustering process also substantially reduces the API calls. More details are in Table \ref{table:cost}.

On the other hand, \name{}'s performance varies significantly based on the underlying model used for instruction selection. While GPT-3.5 yields the best results, open-source models like Llama and Gemma show mixed performance. 
\name{} with Gemma performs reasonably well on MT-Bench but falls short on AlpacaEval2. 
\name{} with Llama, despite its lower MT-Bench score, performs comparably to Alpagasus on AlpacaEval2. 
These results highlight the superiority of \name{} when paired with a strong selection model like GPT-3.5, while also revealing the current limitations of open-source models in selecting unlabeled instructions. Results on the Alpaca are in the appendix.

\noindent\textbf{Cross-dataset generalization.}
Lastly, we analyze how well models trained on one dataset using various sampling techniques generalize to the other dataset.
Our results are presented in Table \ref{table:cross_eval}. 
Here, we made an intriguing observation about models trained on Dolly: the model trained using our approach remarkably performed better than all the baselines by 10\% on the cleaned alpaca test set. 
On the Cleaned Alpaca, Coreset shows comparable performance to \name{} in terms of Cosine Similarity and slightly better in Rouge scores. 
Further, Cleaned Alpaca appears to be a better dataset for cross-evaluation generalization when observing the performance of all baselines trained on it.

\begin{table}[t]
	\begin{center}
	\caption{Win-Tie-Draw from Human evaluation (\%) on Dolly 3k \name{} against Alpagasus on 50 prompts.}
	\label{table:main_humanEval}
    \begin{adjustbox}{width=0.95\linewidth}
	\begin{tabular}{l|ccc}
		\toprule
		Compared Methods & Win & Tie & Lose
		              \\ \midrule
            \name{} vs Alpagasus& \textbf{38} & 27 & 35\\
		\bottomrule
	\end{tabular}
    \end{adjustbox}
    \end{center}
    \vspace{-0.1in}
\end{table}

\begin{table}[t]
	\begin{center}
	\caption{Cross-dataset generalization for 3k sample size. Rouge-L (F1) / Cosine similarity of generated responses from LLaMA-2. The best scores are highlighted in \textbf{bold}, and column names indicate the dataset trained on.}
	\label{table:cross_eval}
    \begin{adjustbox}{width=0.95\linewidth}
	\begin{tabular}{c|cc}
		\toprule
		Methods & Dolly & Cleaned Alpaca
		              \\ \midrule
		Random & {0.205 / 0.589} & 0.260 / 0.669 \\
            OpenEnd & {0.208} / 0.627 & {0.244} / 0.640 \\
            Coreset & {0.208} / 0.651 & \textbf{0.271} / \textbf{0.684}  \\
		\name{} & \textbf{{0.229}} / \textbf{0.668} &  0.263 / 0.683  \\
		\bottomrule
	\end{tabular}
    \end{adjustbox}
    \vspace{-0.2in}
    \end{center}
\end{table}

\begin{table}[t]
	\begin{center}
	\caption{Ablation study. Rouge and cosine similarity of generated responses from fine-tuned LLaMA-2 by selecting 1k examples on Dolly with different methods are compared. The best scores are highlighted in \textbf{bold}.}
	\label{table:ablation}
    \begin{adjustbox}{width=0.95\linewidth}
	\begin{tabular}{cc|cc}
		\toprule
		Division & Selection & Rouge-L (F1) & Cosine Sim
		              \\ \midrule
		Random & Random & {0.239} & {0.589} \\
            Random & OpenEnd & {0.258} & {0.627} \\
            Random & LLM & {0.274} & {0.651} \\
		Sim$_{\tt KMeans}$ & LLM & {0.264} & {0.625} \\ 
  \midrule
            Div$_{\tt KMeans}$ & Random & {0.236} & {0.585} \\
            Div$_{\tt KMeans}$ & OpenEnd & {0.251} & {0.617} \\
		Div$_{\tt KMeans}$ & LLM & \textbf{{0.278}} & \textbf{{0.668}} \\
		\bottomrule
	\end{tabular}
    \end{adjustbox}
    \end{center}
    \vspace{-0.1in}
\end{table}

\begin{table*}[t]
\small
\vspace{-0.1in}
\begin{center}
\caption{Examples of generated sentences from fine-tuned LLMs with different selection methods.}
\label{table:examples}
\begin{adjustbox}{width=0.95\linewidth}
    \begin{tabularx}{\textwidth}{@{}X|X|X@{}}
        \toprule
        \multicolumn{3}{@{}p{\textwidth}@{}}{\textbf{Instruction (\#1)}: Write an email to the head of academic coaching at Services for Students with Disabilities sharing that I've applied to the coaching position. I have experience volunteering with children on the autism spectrum.} \\
        \multicolumn{3}{@{}p{\textwidth}@{}}{\textbf{Ground Truth}: \newline Hi, 
        
        I hope you are having a wonderful day. My name is \_, and I just applied for the Program Staff position at the Services for Students with Disabilities to empower students who have disabilities to reach their academic dreams. My background working in higher education and volunteering with kids on the Autism spectrum combine to bring an empathetic and impactful perspective.
        Please let me know if there is anything else I can do at this time to be considered for this position. 

        With gratitude,} \\
        \midrule
        \textbf{Random} & \textbf{Coreset} & \textbf{Ours} \\
        \textit{\textcolor{red}{Dear Head of Academic Coaching},\newline
            I am writing to share that I have applied to the coaching position. I have experience volunteering with children on the autism spectrum.\newline
            Thank you for your time.\newline
            Sincerely,\newline
            Katie} & \textit{I’m writing to share that I’ve applied to the coaching position. I have experience volunteering with children on the autism spectrum.} &\textit{\textcolor{red}{Dear Head of Academic Coaching}, \newline I am writing to share that I have applied to the coaching position. I have experience volunteering with children on the autism spectrum. \textcolor{blue}{I am passionate about helping children with disabilities and I believe that I would be a great asset to your team. I am confident that I can make a positive impact on the lives of the students that I work with.}
}  \\
        \bottomrule
    \end{tabularx}
    \end{adjustbox}
\end{center}
\vspace{-0.1in}
\end{table*}

\begin{table}[t]
	\begin{center}
	\caption{Cost comparison between \name{} and Alpagasus, a SOTA instruction-tuning data filtering algorithm on Alpaca dataset.}
	\label{table:cost}
    \begin{adjustbox}{width=0.9\linewidth}
	\begin{tabular}{l|ccc}
		\toprule
  		Methods & Size & \# Tokens & Total Cost \\
            \midrule
		Alpagasus & Any & 14M & \$ 23.76 \\
            \midrule
            \multirow{2}{*}{\centering\name{}} & $1k$ & 1.84M & \$ 2.77 \\
             & $3k$ & 1.85M & \$ 2.78 \\
		\bottomrule
	\end{tabular}
    \end{adjustbox}
    \vspace{-0.25in}
    \end{center}
\end{table}

\begin{table}[t]
	\begin{center}
	\caption{Analysis of 1k selected instructions with different approaches on Dolly. Diversity is measured using kNN distance \citep{carbonera2015density} with $k=1$. Perplexity is measured with GPT2-large \citep{radford2019language}, and Length is the number of characters.}
	\label{table:analysis}
    \begin{adjustbox}{width=0.95\linewidth}
	\begin{tabular}{c|ccc}
		\toprule
  		Methods & Diversity ($\uparrow$) & Perplexity ($\downarrow$) & Length \\ \midrule
		Random & 0.721 & 89 & 460 \\
            Coreset & \textbf{0.931} & \underline{47} & \underline{847} \\
            OpenEnd & 0.710 & 71 & 646 \\ 
            \name{} & \underline{0.796} & \textbf{30} & \textbf{1417} \\
		\bottomrule
	\end{tabular}
    \end{adjustbox}
    \vspace{-0.25in}
    \end{center}
\end{table}

\subsection{More Analyses with \name{}}
\noindent\textbf{Ablation study.}
To show the effectiveness of each component of \name{}, we ablate against different combinations of local selection prompting methods (Sec \ref{sec:3.2}) and global division algorithms (Sec \ref{sec:3.3}) on 1k samples from Dolly. 
Table \ref{table:ablation} shows the results. 
We first compare our prompting-based selection method with different non-prompting techniques under the same global division method such as Random and \name{}'s method (called Div$_{\tt Kmeans}$). 
We observe that other selection methods, Random and OpenEndedness, are not as effective in comparison, highlighting the superiority of LLM-based selection in selecting higher-quality instructions for training other LLMs, without true labels.

Next, we compare the global division method, Div$_{\tt Kmeans}$, with Sim$_{\tt Kmeans}$ and random sampling. 
Sim$_{\tt Kmeans}$ clusters the instructions and then constructs input queries with similar instructions rather than diverse ones. 
We observe that Sim$_{\tt Kmeans}$ performs the worst, indicating that having diverse instructions to choose from helps LLMs perform a better local selection. 
This is also highlighted by Random division performing better than Sim$_{\tt Kmeans}$, but worse than Div$_{\tt Kmeans}$.

\noindent\textbf{Human evaluation.} 
Two authors evaluated outputs generated by Alpagasus and \name{} with GPT-3.5 models on 50 prompts (16 from MT-Bench and 34 from AlpacaEval2). The Inter-Annotator Agreement was measured with Krippendorff’s Alpha \citep{hayes2007answering}, yielding a value of 0.52. Results in Table \ref{table:main_humanEval} show \name{} is preferred over the Alpagasus baseline, keeping consistent with the rest of the results.

\noindent\textbf{Comparison of outputs from fine-tuned LLMs.}
We evaluate responses from LLMs fine-tuned with \name{} against Random and Coreset baselines, including one such example on Open QA format in Table \ref{table:examples}, with more in the appendix (Table \ref{table:examples_others}).

\name{} shows a nuanced understanding and response capability. While Random and Coreset give basic, concise answers, \name{} adds personalized, empathetic elements (highlighted in blue), showing deeper instruction comprehension. This reflects the advanced response capabilities needed in instruction-tuned models. 

\begin{figure}[t]
	\centering
	\includegraphics[width=1.0\columnwidth,trim={0cm 0cm 0cm 0cm},clip]{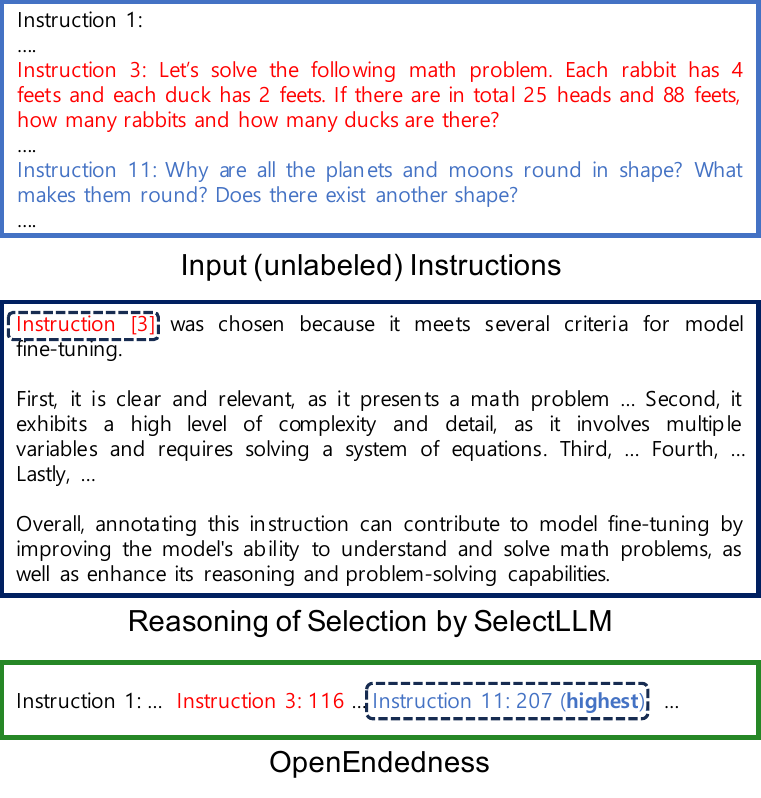}
    \vspace{-0.25in}
    \caption{Qualitative example of selection with a given query composed of 14 instructions on Dolly.}
    \label{fig:llmcot}
    \vspace{-0.2in}
\end{figure}

\noindent\textbf{Analysis of chosen instructions.} 
We provide additional experiments to examine \textit{why \name{} could be effective compared to other selection methods.}
To this end, we first conduct statistical analysis for the instructions selected by Random, Coreset, OpenEnd, and \name{} (Table \ref{table:analysis}).
The results show that (1) \name{} selects high quality (\textit{i.e.}, lower perplexity) instructions with more details (\textit{i.e.}, longer length), and (2) the selected instructions are considerably diverse; it demonstrates the effectiveness of selection by LLMs and composing diverse query via clustering, respectively.  
Next, to further explore the advantage of selection via LLMs over existing approaches, we conduct an additional comparison between \name{} and the method that uses OpenEnd for local selection and Div$_{\tt Kmeans}$ for global division, which is presented in 6th row in Table \ref{table:ablation}. 
Lastly, we present specific examples of the selections with two methods, along with the rationales for the selection with \name{} generated via zero-shot chain-of-thought with ChatGPT \citep{kojima2022large}.
As shown in Figure \ref{fig:llmcot}, we observe the several underlying rationales considered by LLM in making its selection. 
More examples are in Appendix (Figures \ref{fig:llmcotappendix} and \ref{fig:llmcotappendix2}).  

\section{Conclusion}
We introduce \name{}, a new approach that uses LLMs to choose an efficient subset from a set of unlabelled instructions. Our experiments on two popular datasets show that \name{} is more effective than previous selection methods. This demonstrates how LLMs can improve the efficiency of instruction tuning for language models.

\section*{Limitations}
Despite the impressive performance of \name{}, it is not without its limitations. 
A primary concern is the expense associated with utilizing LLMs like ChatGPT for data selection, which can be substantial. 
Additionally, the scalability of \name{} when dealing with exceptionally large datasets, or in scenarios requiring real-time data annotation, remains an area that needs further exploration. 
This aspect is particularly crucial given the ever-increasing size of datasets and the imperative for efficient processing in a wide range of practical applications. 
Hopefully, these limitations will be addressed in the future upon our work.

\section*{Broader Impact and Ethical Implications}

The findings from our research not only establish the proficiency of LLMs in autonomously selecting high-quality data for training but also open new paths for future investigation in this domain. 
The successful application of LLMs in data-constrained environments is demonstrated by the exceptional ability of \name{}. 
This study, therefore, marks a significant stride in the field of instruction tuning for LLMs, paving the way for more efficient and effective training methodologies and expanding the scope of autonomous capabilities of LLMs.
In terms of ethical implications, the potential for any risk is limited to the application of LLMs in our framework, and the general risks associated with them such as LLMs showing bias in selecting certain instructions according to what it believes to be an impactful instruction. Further, bias can also be introduced based on how the prompt is designed by the user, when querying the LLMs in our framework.

\section*{Acknowledgements}
This project has been generously supported by funding from Cisco. We thank the Cisco team and Minnesota NLP group members for their invaluable feedback on both the project and this draft.

\bibliography{custom}
\bibliographystyle{acl_natbib}

\clearpage

\appendix

\section{More Details about Experiments}\label{appendix:more_details}
\subsection{Datasets}
\textbf{1) Dolly \citep{DatabricksBlog2023DollyV2}}: The Dolly dataset, developed by Databricks, Inc., is a comprehensive collection of over 15,000 human-generated instruction-following records. It encompasses a variety of behavioral categories such as brainstorming, classification, closed and open QA, generation, information extraction, and summarization. This dataset was created by Databricks employees and is designed to enhance the interactivity and responsiveness of large language models (LLMs) similar to ChatGPT.

\noindent\textbf{2) Cleaned Alpaca}: Cleaned Alpaca is based on \citep{alpaca} but with responses generated by GPT-4. It addresses several issues found in the original dataset, which was generated using GPT-3. The cleaned dataset incorporates approximately 52,000 instructions.

\subsection{Baselines}
\textbf{1) Random}: As the name suggests, random sampling involves selecting instances from the unlabeled dataset purely at random, without considering their informativeness or representativeness.

\noindent\textbf{2) Cluster-Based Selection (CBS$_{\tt sbert}$ and  CBS$_{\tt inst}$}) \citep{chen2023maybe, su2022one}: Uses clustering for sample selection. CBS$_{\tt sbert}$ involves transforming instructions into vector representations with Sentence-BERT \citep{reimers2019sbert}, whereas the CBS$_{\tt inst}$ method involves transforming the sentences into embeddings using a pre-trained embedder called InstructOR, derived from \citep{su2022one}. InstructOR is supposed to be faster in the conversion of sentences into embeddings. Once we have the embeddings, both these methods have similar steps. We then carry out clustering of the respective embeddings with HDBSCAN \citep{campello2013hdbscan}, followed by selecting samples using the K-Center-Greedy algorithm which focuses on cluster centroids. 

\noindent\textbf{3) Perplexity} \citep{li2023symbolic}: Selects samples based on low per-token perplexity, indicating high model certainty and fluency, akin to the approach in \citep{li2023symbolic}.

\noindent\textbf{4) Diversity} \citep{selfinstruct}: This method utilizes Rouge scores to evaluate the diversity of prompts. For each instruction in the dataset, Rouge scores are computed against a randomly selected subset comprising n samples, where n is strictly less than the total size of the dataset (n < Dataset Size). The selection criterion is based on the aggregation of these Rouge scores. Specifically, we select samples that exhibit the minimum aggregated Rouge scores, thereby ensuring a diverse representation in the final dataset.

\noindent\textbf{5) Open-Endedness} \citep{li2023symbolic}: Determines prompt open-endedness by generating three inferences per prompt, counting unique bigrams between the generated inferences, and selecting samples with the greatest variety of bigrams. This process follows the open-endedness criteria defined for samples with a broader range of chain of thought reasonings in \citep{li2023symbolic}.

\noindent\textbf{6) Coreset} \citep{sener2018active}: Similar to CBS, this method involves transforming data points into their embedding space with Sentence-BERT, then iterates a process to get subsets that maximize the coverage and diversity within each subset with a predetermined subset size. The algorithm selects samples by prioritizing those that maximize the distance to the nearest point already included in the subset, ensuring that the selected samples are diverse within each subset.

\noindent\textbf{7) Length}: Considers the length of prompts (Instruction + Input), focusing on both longer and shorter ones to evaluate their impact. 

\noindent\textbf{8) Alpagasus} \citep{chen2023alpagasus}: Scoring each data point with an auto-grader like ChatGPT based on dimensions such as helpfulness or accuracy and excluding low-quality data.

\subsection{\name{}}

In Algorithms \ref{alg:main} and \ref{alg:kmeans}, we describe the proposed algorithms, presented in Section 3. 
In addition, we present examples of generated sentences from fine-tuned LLMs with different selection methods in Table \ref{table:examples_others}.

\subsection{Implementation details}

Our experiment uses the Dolly and Cleaned Alpaca datasets. 
Depending on the purpose of the experiments, we use the different setups as described below:

\noindent\textbf{Setups for Rouge and Cosine Similarity Evaluations}: 
In Tables \ref{table:main_dolly} and \ref{table:main_alpaca}, we aim to compare the proposed method with various baselines comprehensively. 
To this end, we split the given dataset into training and test ones and evaluated each method. 
Since there were no explicit training or test splits for these datasets, we randomly sampled 1k samples from each dataset to form our test sets. 
This allocation leaves us with 51k samples in the Cleaned Alpaca dataset and 14k samples in the Dolly dataset for training purposes. 
Then, we apply one of our sampling algorithms to each training dataset to select a subset of data, varying the subset size between 1k to 3k samples for training, with an 80:20 training and validation split. 
We fine-tune LLaMA-2 (7B) model \citep{touvron2023llama} by employing QLoRa \citep{dettmers2023qlora}, a model optimization technique, to reduce the memory requirements during the fine-tuning and inference processes. 
For the experiments, we use three different random seeds and then compute the average of the evaluation scores from these three models to derive a final score for each method. We run a total of 20 epochs with a batch size of 6. We use the Paged optimizer and set the gradient accumulation steps at 2. To avoid overfitting and select the best model, we integrate an Early Stopping Callback with a patience of 3 epochs and a threshold of 0.01. Also, for selecting instructions with \name{}, we commonly use ChatGPT (\texttt{gpt-3.5-turbo-0613}).

\noindent\textbf{Setups for MT-Bench and AlpacaEval2 Evaluations}: Next, we compare our method with the current state-of-the-art selection approach, Alpagasus \cite{chen2023alpagasus}. 
For this, we sample from the entire Dolly (15k) and Cleaned Alpaca (52k) datasets, following the paper \cite{chen2023alpagasus}. In addition, we include Random sampling as the baseline. Specifically, we sample 3k from Dolly and 9k from Cleaned Alpaca based on the publicly available sampled data for Alpagasus on both datasets for a direct comparison. We also obtain the entirety of the two datasets from the same source. The sampling techniques for \name{} include ChatGPT versions (\texttt{gpt-3.5-turbo-0125}, \texttt{gpt-4o-0513)}, Llama 3.1-70b-it (Q8), and Gemma-2-27b-it. We fine-tune mistral-7b-v0.3 \citep{jiang2023mistral7b} by employing QLoRa. We use a single fixed random seed for training. We run a total of 3 epochs without Early Stopping. 

\begin{algorithm}[t!]
   \caption{\name{}}
   \label{alg:main}
\begin{algorithmic}
  \State
  \textbf{Input:} Un-annotated instructions $\mathcal{X}$, large language model \texttt{LLM}, input prompt $p_\texttt{sel}$, sentence encoder $g_{\phi}$, number of samples in query $K$, number of queries $T$, number of output $O$
  \State
  \textbf{Output:} Selected indices $S_{\tt all}$
  \vspace{0.05in} 
  \hrule
  \vspace{0.05in} 
  \State \texttt{\color{Gray2} /* Construct input queries */} 
  \State $q_{1}, \cdots, q_{T} \leftarrow \texttt{Diverse-query}(\mathcal{X}, g_{\phi}, K, T)$
  \State $S_{\tt all} \leftarrow \emptyset$ 
  \For{$t=1$ {\bfseries to} $T$}
  \State \texttt{\color{Gray2} /* Selection via LLM */} 
  \State $S_{t} \leftarrow \texttt{LLM}\big(p_{\tt sel}(q_{t}, O)\big)$ 
  \State $S_{\tt all} \leftarrow S_{\tt all} \cup S_{t}$ 
  \EndFor
\end{algorithmic}
\end{algorithm}
\begin{algorithm}[t!]
   \caption{\texttt{Diverse-query}}
   \label{alg:kmeans}
\begin{algorithmic}
  \State
  \textbf{Input:} Un-annotated instructions $\mathcal{X}$, sentence encoder $g_{\phi}$, number of clusters $K$, number of queries $T$
  \State \textbf{Output:} Set of queries $\{q_{t}\}_{t=1}^{T}$
  \vspace{0.05in} 
  \hrule
  \vspace{0.05in} 
  \State $c_{1}, \cdots, c_{K} \leftarrow \texttt{K-means}(\{g_{\phi}(x)\}, K)$
  \State $d_{1,1}, \cdots, d_{N,K} \leftarrow \texttt{l2-dist}(\{g_{\phi}(x)\}, \{c_{k}\})$
  \For{$k=1$ {\bfseries to} $K$}
  \State $I_{k} \leftarrow \texttt{argsort}\{d_{1,k}, \cdots, d_{N,k}\}$
  \EndFor
  \State $A = \{1, \cdots, N\}$
  \For{$t=1$ {\bfseries to} $T$}
  \State $q_{t} \leftarrow \emptyset$ 
  \For{$k=1$ {\bfseries to} $K$}
  \State $s \leftarrow 1$ 
  \While {$|q_{t}| < k$}
    ~~~~\If{$I_{k}(s) \in A$} 
    \State $q_{t} \leftarrow q_{t} \cup \{x_{I_{k}(s)}\}$ 
    \Else
    \State $s \leftarrow s+1$
    \EndIf
    \EndWhile
  \EndFor
  \EndFor
\end{algorithmic}
\end{algorithm}

\section{Comparison with Previous Selection Methods}\label{appendix:discuss_prev}

\begin{table}[t]
\begin{center}
\caption{Comparison of different selection methods.}
\begin{adjustbox}{width=0.95\linewidth}    
\begin{tabular}{c|cccc}
\toprule
\textbf{Sources} & \textbf{Complexity} & \textbf{Information} & \textbf{Flexibility} & \textbf{Cost}\\ \midrule
Input & {\color{blue} Low} & {\color{red} Low} & {\color{red} Low} & {\color{blue} Low}\\ 
Response & {\color{red} High} & {\color{blue} High} &  {\color{red} Low} & Medium\\ 
LLMs & Medium & {\color{blue} High} & {\color{blue} High} & {\color{red} High}\\ \bottomrule
\end{tabular}
}
\label{table:comparison}
\vspace{-0.2in}
    \end{adjustbox}
    \end{center}
\end{table}

In this section, we provide a detailed comparison between \name{} and previous approaches for sample selection.
First, we divide the existing sample selection approaches into two categories: \textit{input-based} and \textit{response-based} ones. 
Input-based approaches only use the input text to select samples, \textit{e.g.}, given instruction without the corresponding label. 
For example, \citet{chen2023maybe} transforms input instruction into embedding space, and then applies clustering and K-Center-Greedy algorithms.
In contrast, response-based approaches first generate responses with the external model, and then select samples using both instruction and artificial response; for instance, one can utilize the fine-tuned LLMs with small labeled instructions \citep{kung2023active} or fixed pre-trained LLMs \citep{wang2022self}.

Since the two approaches rely on different sources to extract the information for the samples, they have distinct characteristics. 
First, in terms of the complexity of the method, an input-based one is much simpler than a response-based one as it does not require additional processes like model fine-tuning or generation of the response.
However, response-based one can utilize more information about the sample, thanks to the generated response, while it requires more cost to obtain that. 
In the case of \name{}, it's not very complex as the user can easily select the samples using the LLMs. 
Also, \name{} utilizes extensive information within LLMs while inferring the importance of each sample to select the important ones with prompting. 
Although it requires the cost for prompting with ChatGPT-based approaches in our experiments, we remark that \name{} exhibits the unique capability that could be flexibly adapted for the desired property and is still much cheaper than its counterparts like Alpagasus.
We summarize the comparison of different approaches in Table \ref{table:comparison}.

\section{Verifying Capability of LLMs for Selecting Unlabelled Instruction}\label{appendix:list_prompt}

In this section, we present the full input prompt to verify whether LLMs could infer the importance of unlabelled instructions, providing details for the experiments in Figure \ref{fig:figure_intuition}.
We adapted the prompt from the recent work using LLMs for text re-ranking \citep{sun2023chatgpt}, and the prompt is presented in Figure \ref{fig:prompt_list}.

\begin{figure}[t!]
    \centering
    \begin{lstlisting}
This is RankGPT, an intelligent assistant that can rank instructions based on their impactfulness and informativeness for model fine-tuning, when labeled by humans, like active learning.

The following are @{num}@ examples of instructions that describe a task, each indicated by a number identifier [].
$
[1]
### Instruction: {Example #1 Instruction}
### Input: {Example #1 Input}
.
.
.
[N]
### Instruction: {Example #N Instruction}
### Input: {Example #N Input}
$
I will rank the @{num}@ instructions above based on their impactfulness and informativeness for model fine-tuning when labeled by humans, like active learning. The examples will be listed in descending order using identifiers, and the most impactful examples should be listed first, and the output format should be [] > [] > etc, e.g., [1] > [2] > etc.

The ranking results of the @{num}@ examples (only identifiers) is
    
    \end{lstlisting}
    \caption{Full prompt to investigate the capability of LLM to infer the importance of instructions without labels.}
    \label{fig:prompt_list}
    \vspace{-0.1in}
\end{figure}

\begin{table*}[t]
\small
\vspace{-0.1in}
\begin{center}
\caption{More instances of closed QA showing \name{} giving more concise answers. For example, in Instruction \#3, all three
models arrive at the correct answer, however, only \name{} clearly distinguishes between the two
terms by providing detailed definitions, highlighting its capacity to deliver more informative and contextually rich explanations}
\label{table:examples_others}
\begin{adjustbox}{width=1\linewidth}
    \begin{tabularx}{\textwidth}{@{}X|X|X@{}}
        \toprule
        \multicolumn{3}{@{}p{\textwidth}@{}}{\textbf{Instruction (\#2)}: Where was Sarah Schechter born?} \\
        \multicolumn{3}{@{}p{\textwidth}@{}}{\textbf{Input}: Schechter was born in Boston and grew up in Brookline, Massachusetts and New York City. Her father, Danny Schechter, a grandson of Russian-Jewish immigrant socialists, was a human rights activist, author, filmmaker and television producer.} \\
        \multicolumn{3}{@{}p{\textwidth}@{}}{\textbf{Ground Truth}: Sarah Schechter was born in Boston, Massachusetts.} \\
        \midrule
        \textbf{Random} & \textbf{Coreset} & \textbf{Ours} \\
        \textit{Sarah Schechter was born in Boston and grew up in Brookline, Massachusetts and New York City. Her father, Danny Schechter, a grandson of Russian-Jewish immigrant socialists, was a human rights activist, author, filmmaker and television producer.\newline} & \textit{Sarah Schechter was born in Boston and grew up in Brookline, Massachusetts and New York City. Her father, Danny Schechter, a grandson of Russian-Jewish immigrant socialists, was a human rights activist, author, filmmaker and television producer.} &\textit{Sarah Schechter was born in Boston, Massachusetts.
} \\
    \end{tabularx}
    \end{adjustbox}
\begin{adjustbox}{width=1.0\linewidth}
    \begin{tabularx}{\textwidth}{@{}X|X|X@{}}
        \toprule
        \multicolumn{3}{@{}p{\textwidth}@{}}{\textbf{Instruction (\#3)}: Which is a species of fish? Red snapper or Red Sox} \\
        \multicolumn{3}{@{}p{\textwidth}@{}}{\textbf{Ground Truth}: Red snapper} \\
        \midrule
        \textbf{Random} & \textbf{Coreset} & \textbf{Ours} \\
        \textit{Red snapper} & \textit{Red snapper is a species of fish.} &\textit{Red snapper is a species of fish.\newline
Red Sox is a baseball team.
} \\
        \bottomrule
    \end{tabularx}
    \end{adjustbox}
\end{center}
\end{table*}

\begin{table*}[t]
    \caption{Experimental results on Dolly \citep{DatabricksBlog2023DollyV2}. Rouge-L (F1) and Cosine similarity of generated responses from fine-tuned LLaMA-2 models on the entire dolly dataset are compared with a model trained on only 3k samples chosen by \name{}.} 
    \vspace{-0.1in}
	\begin{center}
	\begin{adjustbox}{width=0.60\linewidth}
	\begin{tabular}{r|c|cc}
 		\toprule
		Methods & Rouge-L (F1) & Cosine Similarity\\ \midrule
  Dolly-14k
  & 0.292
  & 0.689 \\ \midrule
  \name{}-3k
  & 0.289 
  & 0.686 \\ 
 		\bottomrule
	\end{tabular}
    \end{adjustbox}
    \end{center}
    \vspace{-0.1in}
    \label{table:main_full_dolly}
\end{table*}
\begin{table*}[t]
    \caption{Experimental results on Cleaned Alpaca \citep{DatabricksBlog2023DollyV2}. Rouge-L (F1) and Cosine similarity of generated responses from fine-tuned LLaMA-2 models on the entire Cleaned Alpaca dataset is compared with a model trained on only 3k samples chosen by \name{}.} 
    \vspace{-0.1in}
	\begin{center}
	\begin{adjustbox}{width=0.60\linewidth}
	\begin{tabular}{r|c|cc}
 		\toprule
		Methods & Rouge-L (F1) & Cosine Similarity\\ \midrule
  CleanedAlpaca-51k
  & 0.356
  & 0.755 \\ \midrule
  \name{}-3k
  & 0.301
  & 0.707 \\ 
 		\bottomrule
	\end{tabular}
    \end{adjustbox}
    \end{center}
    \vspace{-0.1in}
    \label{table:main_full_alpaca}
\end{table*}

\section{More Analyses with Chosen Instructions}\label{appendix:more_analyses}

\begin{figure}[t!]
    \centering
    \begin{lstlisting}
The following are @{N}@ candidate instructions that describe a task, each indicated by a number identifier [].
$
[1]
### Instruction: {Example #1 Instruction}
### Input: {Example #1 Input}
.
.
.
[N]
### Instruction: {Example #N Instruction}
### Input: {Example #N Input}
$
Examine the provided list of @{N}@ instructions, each uniquely identified by a number in brackets []. 

Your task is to select @{num}@ instructions that excel in various aspects.

Look for instructions that are clear and relevant, exhibit a high level of complexity and detail, represent a diverse range of scenarios and contexts, offer significant instructional value and potential learning gain, and present unique challenges and specificity.

These selected instructions should ideally be the most beneficial for model fine-tuning after being annotated by human annotators.
       
Present your selections using the format []. e.g., [1,2] or [2,3].

The most impactful @{num}@ instructions (only identi-fiers) are: ${prev_selection}$

Explain why it was chosen, focusing on how it meets the above criteria and its potential contribution to model fine-tuning. Rationale for selection:
    
    \end{lstlisting}
    \caption{Prompt to generate reasoning for the selection.}
    \label{fig:prompt_select}
    \vspace{-0.1in}
\end{figure}
\begin{figure}[t!]
    \centering
    \begin{lstlisting}
Question: Given the following responses to the target question, determine which is more informative and plausible to answer a given question properly.

Response 1:
${Method #1 response}$

Response 2:
${Method #2 response}$

Target Question:
${question}$

Your Task:
Identify which response (Response 1 or Response 2) is more informative and plausible to answer a given question at hand. Choices: [Response 1, Response 2]. Answer with less than 3 words.

Answer:
    \end{lstlisting}
    \caption{Prompt for GPT-4 evaluation on \name{} against Random and Coreset. \{\textcolor{blue}{blues}\} indicate the place for the inputs. To prevent order bias of LLMs, we ask GPT-4 twice with changed order of responses.}
    \label{fig:prompt_gpt4}
    \vspace{-0.1in}
\end{figure}

\noindent In this section, we provide more details about additional analyses of chosen instructions. 
First, in Figure \ref{fig:prompt_select}, we present a full prompt to generate reasoning for the selection by \name{}, which is used in Figure \ref{fig:llmcot}. 
We analyze 10 clusters, each containing 14 instructions, unveiling the LLM's intricate selection criteria. 
Key factors influencing the LLM's choices include clarity and relevance, complexity and detail, and the potential for instructional value and learning gains. 
The LLM shows a propensity for instructions that require a nuanced understanding and provide substantial learning opportunities, such as querying specific information about diverse topics like the Lollapalooza music festival, process mining, and the top speed of a Kia Stinger. 
Moreover, LLM consistently selects instructions that pose unique challenges and demand specificity, thereby testing and expanding its knowledge across various domains. Figures 9 and 10 show more such detailed examples with the rationales provided by the LLM for its selection compared against the selection by OpenEndedness. 

\section{Comparison of Selected Sample with the Entire Dataset}\label{appendix:discuss_entiredataset}

We compared the \name{}-3k sampled model with models trained on full Dolly and Cleaned Alpaca datasets.
As detailed in \ref{table:main_full_dolly} and \ref{table:main_full_alpaca}, \name{}-3k nearly equals the full Dolly dataset model in performance and achieves 85\% of the full dataset's performance on the Cleaned Alpaca dataset, as per the Rouge score. This underscores \name{}'s efficiency in filtering out noise from the entire dataset, especially in the case of Dolly.  While our results are akin to the data filtering approach in \citep{chen2023alpagasus}, which also samples 3k instructions from the Dolly dataset, there are notable differences. Their method outperforms the full dataset, possibly due to their inclusion of sample outputs in the filtering process, an aspect not considered in our approach.

\section{GPT-Evaluation of models}\label{appendix:discuss_gpteval}
We evaluate the quality of generated responses between LLaMA-2 finetuned with \name{} and other methods on a 1k size instruction dataset, randomly sampled from the Dolly dataset. 
We ask GPT-4 to choose the better response to a given instruction, similar to \citet{liu2023gpteval}. 
As shown in Table \ref{table:main_gpt_eval}, \name{} wins in 52\% of the cases when compared to Random sampling and 44\% cases when compared to Coreset sampling, further showcasing better inference quality of \name{}. The prompt for the evaluation is Figure \ref{fig:prompt_gpt4}.

\section{\name{} with GPT-4o}\label{appendix:discuss_gpt4o}

We ran \name{} with GPT-4o-0125, and the results are provided in Table \ref{table:main_benchmark_gpt4}. 
We have a surprising observation in GPT-4o performing worse than GPT 3.5 based \name{}. 
While we need to do more analysis on this, one explanation we have is that experimented mainly with GPT 3.5 and then extended the framework to other LLMs.
The prompt was optimized for GPT 3.5, and in the future, we would like to see if other factors contribute to such a performance as well. 

\section{MT-Bench and AlpacaEval2 on Cleaned Alpaca}\label{appendix:discuss_mtbench_alpaca}

Our MT-Bench and AlpacaEval2 results on the Alpaca dataset with 9k samples are provided in Table \ref{table:main_benchmark_alpaca}. Our method performs the best on Length Controlled Win rate in AlpacaEval2, however, is outperformed by Alpagasus on MT-Bench. 
Random baseline, surprisingly, also performs really well.

\section{Detailed MT-Bench Results}\label{appendix:discuss_mtbench}
We also look at the detailed MT-Bench results by each subject or topic tested in this particular benchmark. The result between the best performing \name{} model with GPT 3.5 and the baselines on Dolly dataset is provided in Table \ref{table:main_benchmark_extended}.

\begin{table}[t]
	\begin{center}
	\caption{Win-Tie-Draw from GPT-4 evaluation (\%) on \name{} against Random and Coreset with Dolly.}
	\label{table:main_gpt_eval}
    \begin{adjustbox}{width=0.95\linewidth}
	\begin{tabular}{c|ccc}
		\toprule
		Compared Methods & Win & Tie & Lose
		              \\ \midrule
		  \name{} vs Random & \textbf{52.1} & 18.4 & 29.5 \\
            \name{} vs Coreset& \textbf{44.2} & 25.4 & 30.4\\
		\bottomrule
	\end{tabular}
    \end{adjustbox}
    \end{center}
    \vspace{-0.1in}
\end{table}

\begin{table*}[t]
\small
\vspace{-0.1in}
\begin{center}
    \caption{Category-wise MT-Bench evaluation results of fine-tuned Mistral-7B-inst-v0.3 on $3k$ samples of Alpaca dataset selected with different methods. The MT-Bench scores range from 1 to 10. The highest scores are \textbf{bolded}.}
    \vspace{-0.1in}
	\begin{center}
	\begin{adjustbox}{width=1\textwidth}
	\begin{tabular}{l|cccccccc|c}
            \toprule
            \textbf{Model} & \textbf{Writing} & \textbf{Roleplay} & \textbf{Reasoning} & \textbf{Math} & \textbf{Coding} & \textbf{Extraction} & \textbf{STEM} & \textbf{Humanities} & \textbf{Overall} \\
            \midrule
            \midrule
            Mistral 7B v0.3 & 2.85 & 2.65 & 1.60 & 1.10 & 2.20 & 2.18 & 3.50 & 2.75 & 2.33 \\
            Random & 7.05 & 7.00 & \textbf{5.75} & \textbf{2.55} & 3.30 & 5.95 & 6.00 & 8.95 & 5.44 \\
            Alpagasus & \textbf{7.58} & 6.65 & 4.25 & 1.85 & 2.40 & \textbf{6.45} & 7.35 & 9.00 & 5.69 \\
            SelectLLM GPT 3.5 & 7.20 & \textbf{7.05} & 5.40 & 2.30 & \textbf{4.20} & 6.30 & \textbf{7.38} & \textbf{9.10} & \textbf{6.12} \\
            \bottomrule
    \end{tabular}
    \end{adjustbox}
    \end{center}
    \vspace{-0.1in}
    \label{table:main_benchmark_extended}
\end{center}
\end{table*}
\begin{table}[t]
    \caption{MT-Bench and AlpacaEval2 evaluation results of fine-tuned Mistral-7B-inst-v0.3 on $3k$ samples of Dolly dataset selected with different methods. The MT-Bench scores range from 1 to 10, while AlpacaEval2 metrics include Length-Controlled (LC) win rates and raw win rates (WR) as percentages. The highest scores are \textbf{bolded}.}
    \vspace{-0.1in}
	\begin{center}
	\begin{adjustbox}{width=1\linewidth}
	\begin{tabular}{l|c|c|c>{\centering\arraybackslash}p{2em}|>{\centering\arraybackslash}p{2em}}
            \toprule
            \multirow{2}{*}{Model} & MT-Bench & \multicolumn{2}{c}{AlpacaEval2 (\%)} \\
            & (1-10) & LC & WR \\
            \midrule
            Mistral-7B (vanilla) &  2.33 & - & - \\
            Random & 5.44 & 4.00 & 1.93 \\
            Alpagasus & 5.69 & 4.76 & 2.48 \\ \midrule
            SLLM GPT 3.5 & \textbf{6.12} & \textbf{6.72} & \textbf{3.29} \\
            SLLM GPT 4 & 5.99 & 5.57 & 2.73 \\
            SLLM Llama & 3.36 & 4.75 & 2.3 \\
            SLLM Gemma & 5.00 & 4.32 & 2.11 \\
            \bottomrule
    \end{tabular}
    \end{adjustbox}
    \end{center}
    \vspace{-0.1in}
    \label{table:main_benchmark_gpt4}
\end{table}
\begin{table}[t]
    \caption{MT-Bench and AlpacaEval2 evaluation results of fine-tuned Mistral-7B-inst-v0.3 on $9k$ samples of Alpaca dataset selected with different methods. The MT-Bench scores range from 1 to 10, while AlpacaEval2 metrics include Length-Controlled (LC) win rates and raw win rates (WR) as percentages. The highest scores are \textbf{bolded}.}
    \vspace{-0.1in}
	\begin{center}
	\begin{adjustbox}{width=1\linewidth}
	\begin{tabular}{l|c|c|c>{\centering\arraybackslash}p{2em}|>{\centering\arraybackslash}p{2em}}
            \toprule
            \multirow{2}{*}{Model} & MT-Bench & \multicolumn{2}{c}{AlpacaEval2 (\%)} \\
            & (1-10) & LC & WR \\
            \midrule
            Mistral-7B (vanilla) &  2.33 & - & - \\
            Random & 5.82 & 8.44 & \textbf{3.73} \\
            Alpagasus & \textbf{5.84} & 7.43 & 3.23 \\ \midrule
            SLLM GPT 3.5 & 5.52 & \textbf{8.6} & 3.42 \\
            SLLM GPT 4 & 5.42 & 7.72 & 3.17 \\
            SLLM Llama & 5.45 & 7.44 & 3.04 \\
            SLLM Gemma & 5.49 & 5.85 & 2.48 \\
            \bottomrule
    \end{tabular}
    \end{adjustbox}
    \end{center}
    \vspace{-0.1in}
    \label{table:main_benchmark_alpaca}
\end{table}

\begin{figure*}[t]
	\centering
	\includegraphics[width=1.0\textwidth,trim={0cm 0cm 0cm 0cm},clip]{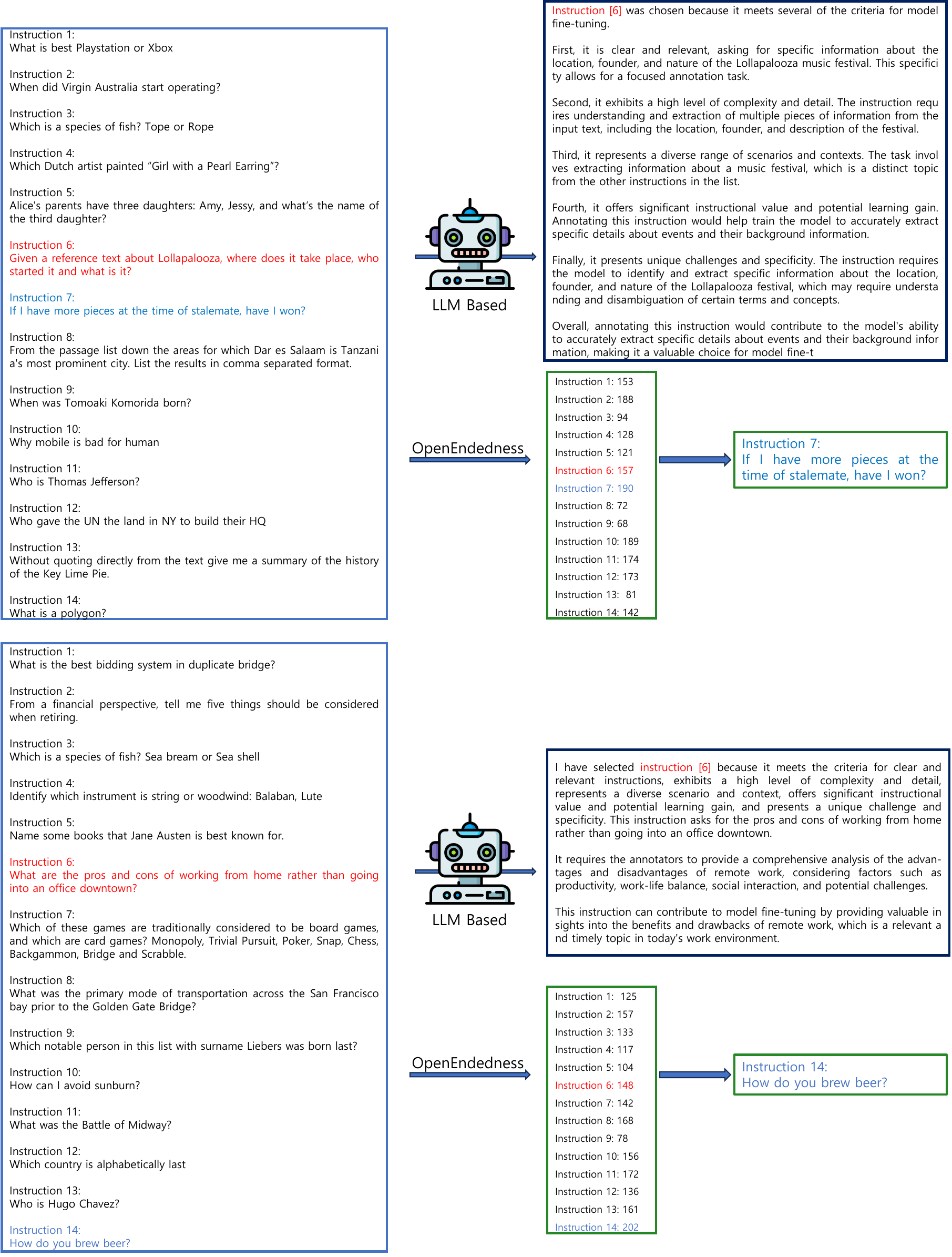}
    \caption{Selection of an Instruction from a given cluster using LLM Based prompting (Red) along with its Chain of Thought Reasoning compared to selection based on the OpenEndedness scores of the given instructions (Blue).}
    \label{fig:llmcotappendix}
    \vspace{-0.1in}
\end{figure*}
\begin{figure*}[t]
	\centering
	\includegraphics[width=1.0\textwidth,trim={0cm 0cm 0cm 0cm},clip]{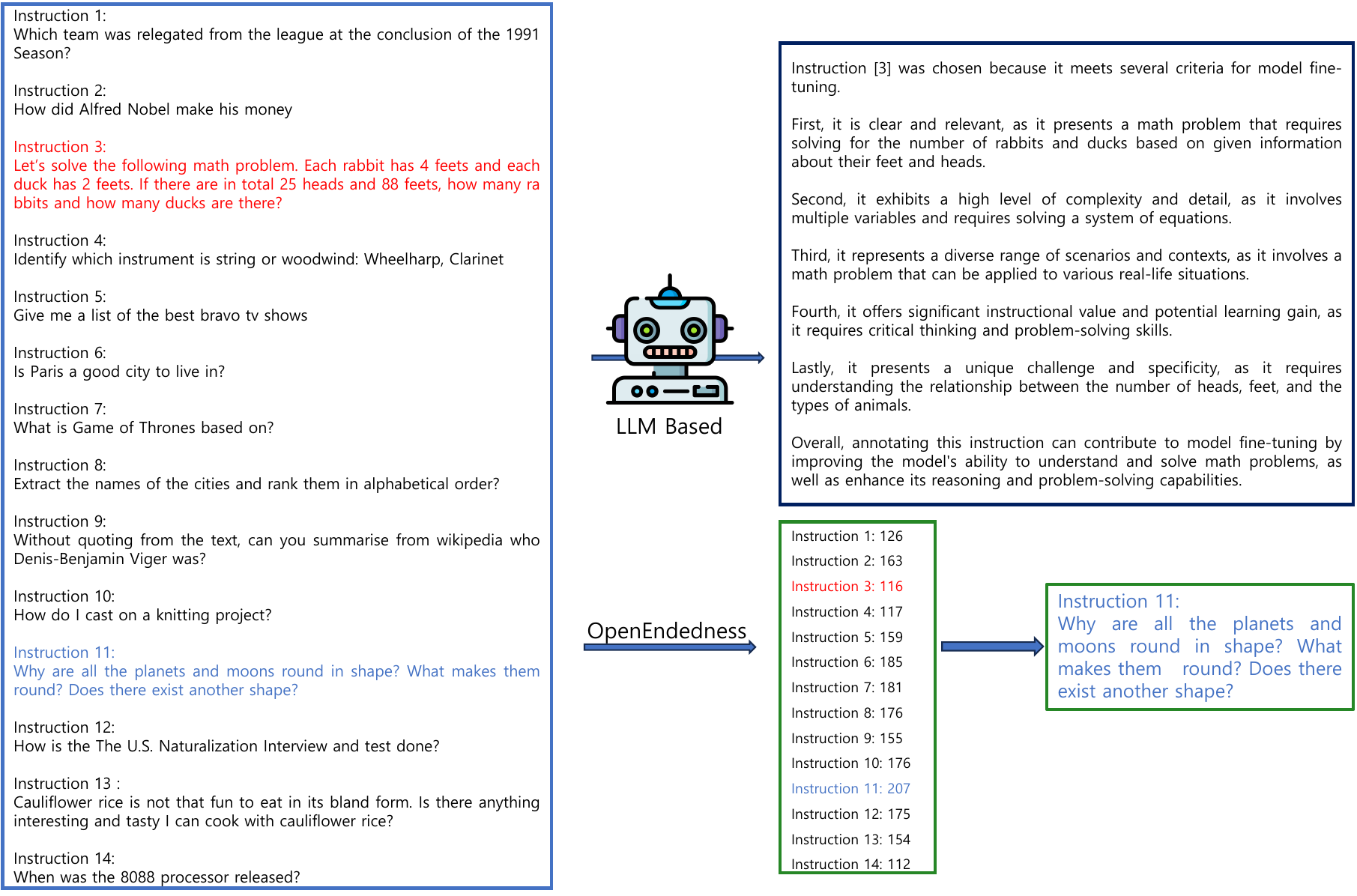}
    \caption{Selection of an Instruction from a given cluster using LLM Based prompting (Red) along with its Chain of Thought Reasoning compared to selection based on the OpenEndedness scores of the given instructions (Blue). This is based on the example shown in the main paper.}
    \label{fig:llmcotappendix2}
    \vspace{-0.1in}
\end{figure*}

\end{document}